\documentclass[preprint,12pt]{elsarticle}

\usepackage{amssymb}
\usepackage{longtable}
\usepackage[utf8]{inputenc} % allow utf-8 input
\usepackage[T1]{fontenc}    % use 8-bit T1 fonts
\usepackage{url}            % simple URL typesetting
\usepackage{booktabs}       % professional-quality tables
\usepackage{amsfonts}       % blackboard math symbols
\usepackage{nicefrac}       % compact symbols for 1/2, etc.
\usepackage{microtype}      % microtypography
\usepackage{graphicx}
\usepackage{amsmath}
\usepackage[table]{xcolor}
\usepackage[warn]{textcomp}
\usepackage{epsfig}
\usepackage{verbatim}
\usepackage{multirow}
\usepackage{txfonts}
\usepackage{lipsum}
\usepackage{float}
\usepackage{lineno}
\usepackage[utf8]{inputenc}
\usepackage[T1]{fontenc}
\usepackage{fourier, erewhon}
\usepackage{geometry}
\usepackage{array, caption, tabularx, makecell, booktabs}
\setcellgapes{3pt}

\journal{Journal of XXXX}

\begin{document}

\begin{frontmatter}


\title{Artificial Intelligence Enhanced Rapid and Efficient Diagnosis of Mycoplasma Pneumoniae Pneumonia in Children Patients}

%% use the corref command within \author for corresponding author footnotes;s

\renewcommand{\thefootnote}{\fnsymbol{footnote}}

\author{Chenglin Pan$^{1}$, Kuan Yan$^{2}$, Xiao Liu$^{3}$, Yanjie Chen$^{1}$, Yanyan Luo$^{4}$, Xiaoming Li$^{5}$, Zhenguo Nie$^{6}$\footnote[1]{Address all correspondence to these authors (email: zhenguonie@tsinghua.edu.cn; xinjunliu@tsinghua.edu.cn).}, Xinjun Liu$^{6*}$}

\address{
$^{1}$Department of Pediatrics, Shanghai Tenth People’s Hospital, Tongji University School of Medicine, Shanghai 200072, China\\
$^{2}$Department of Software Engineering, Xiamen University, Fujian 361105, China\\
$^{3}$School of Electrical and Information Engineering, The University of Sydney, Camperdown NSW 2006, Australia\\
$^{4}$Department of Pediatrics, Huai'an First People's Hospital, Nanjing Medical Universiyt, Huai'an, Jiangsu 223300, China\\
$^{5}$Department of Pediatrics, Hainan Maternal and Children's Medical Center, Haikou, Hainan 570201, China\\
$^{6}$Department of Mechanical Engineering, Tsinghua University, Beijing 100084, China\\
}

\begin{abstract}
Artificial intelligence methods have been increasingly turning into a potentially powerful tool in the diagnosis and management of diseases. In this study, we utilized logistic regression (LR), decision tree (DT), gradient boosted decision tree (GBDT), support vector machine (SVM), and multilayer perceptron (MLP) as machine learning models to rapidly diagnose the mycoplasma pneumoniae pneumonia (MPP) in children patients. The classification task was carried out after applying the preprocessing procedure to the MPP dataset. The most efficient results are obtained by GBDT. It provides the best performance with an accuracy of 93.7\%. In contrast to standard raw feature weighting, the feature importance takes the underlying correlation structure of the features into account. The most crucial feature of GBDT is the “pulmonary infiltrates range” with a score of 0.5925, followed by “cough” (0.0953) and “pleural effusion” (0.0492). We publicly share our full implementation with the dataset and trained models at \url{https://github.com/zhenguonie/2021_AI4MPP}.
\end{abstract}


\end{frontmatter}

%\linenumbers

\section{Introduction}
\label{S:1}
Mycoplasma Pneumoniae Pneumonia (MPP) is one of the major pathogens in respiratory tract infections in children and young adults, manifesting from asymptomatic infection to potential fatal pneumonia. MPP accounts for 10\%-40\% community-acquired pneumonia (CAP) of school-aged children and adolescents \cite{biondi2014treatment, bradley2011management, korppi2004incidence}. MPP infections show an endemic transmission pattern with cyclic epidemics every 3-5 years \cite{atkinsontp2008waiteskb, lind1997seroepidemiological}, which increases the rate of morbidity, mortality, as well as the cost of healthcare in society. Although most MPP infections in children are known as mild and self-limiting, some cases need hospitalization, even in rare cases, MPP can cause extrapulmonary manifestations, including neurologic, dermatologic, hematologic and cardiac syndromes which can result in hospitalization and death \cite{olson2015outbreak, tsiodras2005central}. Macrolide antibiotics are commonly used drugs for the treatment of MPP infection. With the widespread or inappropriate use of antibiotics, and has become an emerging threat worldwide \cite{diaz2015investigations, dumke2010occurrence, pereyre2013increased}, especially in Asia in recent years \cite{ho2015emergence, peirano2014characteristics, tanaka2017macrolide}.

Artificial intelligence methods have emerged as a potentially powerful tool to aid in diagnosis and management of diseases, mimicking and perhaps even augmenting the clinical decision-making of human physicians \cite{perer2015mining}. Due to the high infection rate and severe sequelae of MPP in children patients, there may be a crucial role for AI approaches for the rapid diagnosis based on the basic routine inspections, including demographics and clinical presentations. AI-based systems, which can assist in the diagnosis of MPP as precisely under the epidemic and emergency of MMP, precision can decrease macrolide-resistant mycoplasma pneumoniae (MRMP) \cite{zech2018variable,ge2019predicting,stephen2019efficient,tougaccar2020deep,jakhar2018big}, on the other aspect, it is convenient in some areas, especially in which few healthcare providers in rural China. However, to the best of our knowledge, there is little research in AI based rapid diagnosis on MPP in children patients.

Firstly, we implement five machine learning based classifiers, including logistic regression (LR), decision tree (DT), gradient boosted decision tree(GBDT), support vector machine (SVM), and multilayer perceptron (MLP), in the rapid diagnosis. We collect the training data in multi-center inpatient departments of china. All five AI classifiers are trained and validated on the dataset. The result shows that GBDT produced the best results with an overall accuracy rate of 0.937, and the decision tree came out as the second best with an overall accuracy of 0.884, followed by MLP and logistic regression with overall accuracy rates of 0.863 and 0.695 respective. Besides, feature importance analysis indicates the pulmonary infiltrates range plays a predominant role among all the 42 symptom features with a weight of 0.5925.

\section{Related Work}
\label{S:2}
Our review focuses on studies that highlight AI-aided pneumonia diagnosis and five machine learning frameworks closely related to our work.

\subsection{Artificial Intelligence for pneumonia diagnosis}
Pneumonia is an important infectious disease in the world, which is related to high morbidity and mortality. In 2019, data from the World Health Organization showed that pneumonia is the biggest cause of death from infectious diseases in children.

Over the past decades, more and more researchers were using artificial intelligence technology to diagnose pneumonia. Integrating artificial intelligence into pneumonia diagnosis has become a trend. We use "Pneumonia" as the subject term to search on the Web of Science database with the topic related to “Artificial Intelligence” from January 1, 2006 to October 23, 2020. We analyze the scientific production by country during the past fifteen years. 

\begin{figure}[!t]
\centering
\includegraphics[width=0.7\linewidth]{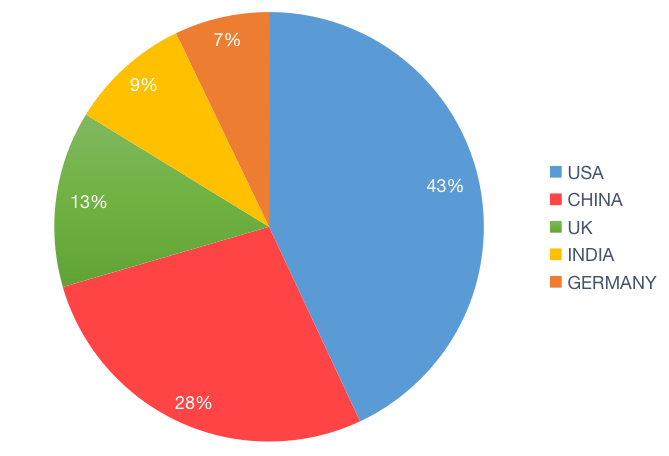}
\caption{The proportion of the top 5 countries ranked on research publications in AI-aided pneumonia diagnosis (2006-2020)}
\label{proportion_comparison}
\end{figure}

\begin{figure}[ht]
\centering\includegraphics[width=0.8\linewidth]{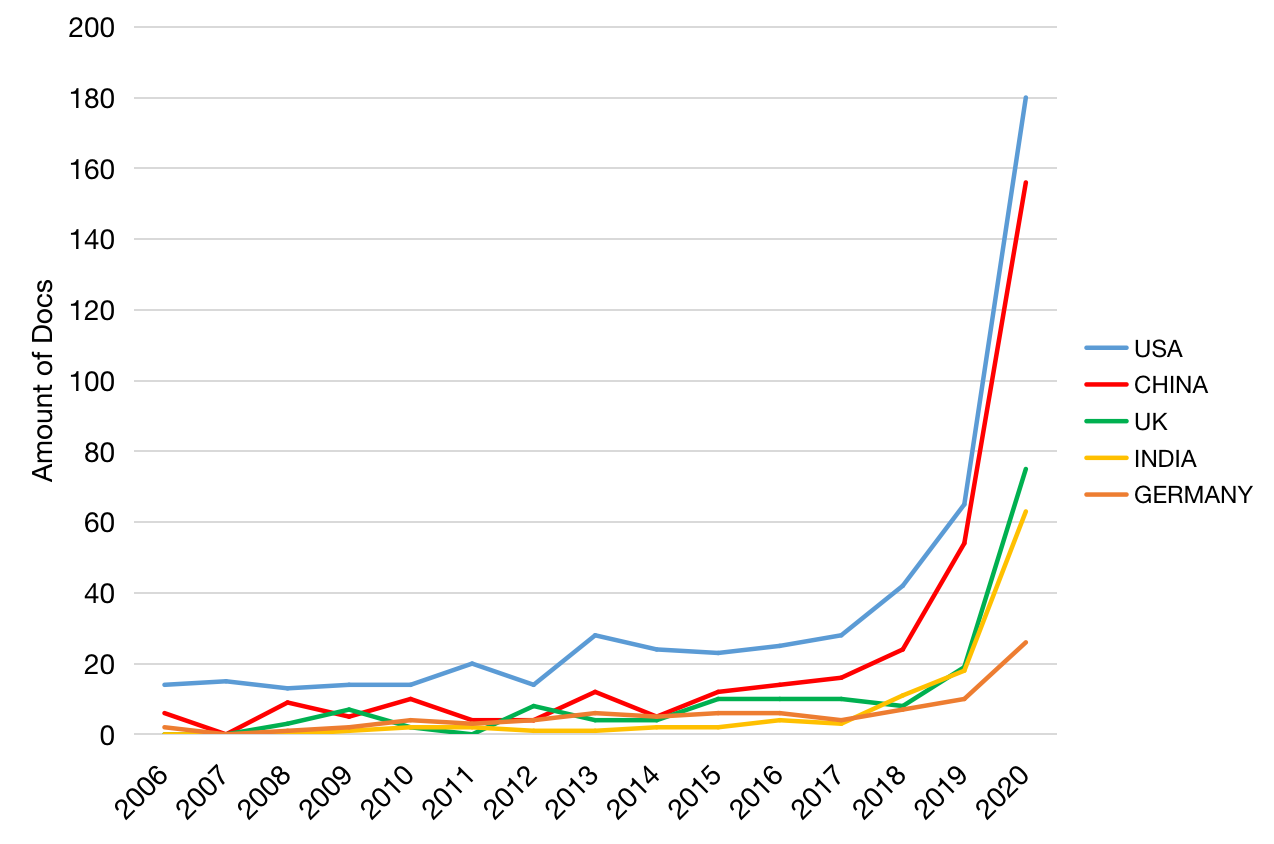}
\caption{The growth curves of the top 5 countries ranked by the number of publications from 2006 to 2020}
\label{linear_growth_model}
\end{figure}

A total of 1,851 publications are received. The scientific production in this research field has increased massively in the past years, from only 33 publications in 2006 to 833 in 2020. USA, China, UK, India, and Germany are the top five most productive countries. Figure \ref{proportion_comparison} shows the proportion of these five countries, and Figure \ref{linear_growth_model} shows their growth over the past fifteen years.

USA is the most productive country with a proportion of 28.039\%, then followed by China (17.882\%), UK (8.644\%), India (5.943\%) and Germany (4.646\%). As shown in Table \ref{table1}, Table \ref{table2}, Table \ref{table3} and Table\ref{table4}. We list the top 10 countries with the most publications in every five years from 2006 to 2020. It also shows the proportion and evolution over time.
\begin{table}[ht!]
  \centering
  \caption{Publications in total}
  \begin{tabular}{l |l| l |l }
   \hline
    \toprule
    %\multicolumn{2}{c}{Part}                   
    \cmidrule(r){1-4}
     \textbf{Country}     & \textbf{No. of docs} & \textbf{\%}    &\textbf{ PPD}  \\
    \hline\hline
    %\midrule
   USA         & 519       & 28.039 & -8.486 \\
China       & 331       & 17.882 & 5.291  \\
UK          & 160       & 8.644  & 3.353  \\
India       & 110       & 5.943  & 6.009  \\
Germany     & 86        & 4.646  & -0.39  \\
Canada      & 78        & 4.214  & 1.677  \\
Italy       & 75        & 4.052  & -1.401 \\
France      & 67        & 3.62   & -3.062 \\
Netherlands & 60        & 3.241  & -2.161 \\
Australia   & 58        & 3.133  & 1.219 \\
    \hline
  \end{tabular}
  \label{table1}
\end{table}

\begin{table}[ht!]
  \centering
  \caption{Publications in 2006-2010}
  \begin{tabular}{l |l| l  }
   \hline
    \toprule
    %\multicolumn{2}{c}{Part}                   
    \cmidrule(r){1-3}
     \textbf{Country}     & \textbf{No. of docs} & \textbf{\%}  \\
    \hline\hline
    %\midrule
USA         & 70        & 34.146 \\
China       & 30        & 14.634 \\
Spain       & 13        & 6.341  \\
France      & 12        & 5.854  \\
Italy       & 12        & 5.854  \\
UK          & 12        & 5.854  \\
Japan       & 11        & 5.366  \\
Netherlands & 10        & 4.878  \\
Germany     & 9         & 4.39   \\
Canada      & 6         & 2.927 \\
    \hline
  
   %\hline
  \end{tabular}
  
  \label{table2}
\end{table}

\begin{table}[ht!]
  \centering
  \caption{Publications in 2011-2015}
  \begin{tabular}{l |l| l  }
   \hline
    \toprule
    %\multicolumn{2}{c}{Part}                   
    \cmidrule(r){1-3}
     \textbf{Country}     & \textbf{No. of docs} & \textbf{\%}  \\
\hline\hline
     USA         & 109       & 33.956 \\
China       & 37        & 11.526 \\
UK          & 26        & 8.1    \\
Germany     & 24        & 7.477  \\
France      & 18        & 5.607  \\
Netherlands & 14        & 4.361  \\
Australia   & 12        & 3.738  \\
Canada      & 11        & 3.427  \\
Switzerland & 11        & 3.427  \\
Belgium     & 10        & 3.115 \\
    \hline\hline
    %\midrule

    \hline
  
   %\hline
  \end{tabular}
  
  \label{table3}
\end{table}

\begin{table}[H]
  \centering
  \caption{Publications in 2016-2020 }
  \begin{tabular}{l |l| l  }
   \hline
    \toprule
    %\multicolumn{2}{c}{Part}                   
    \cmidrule(r){1-3}
     \textbf{Country}     & \textbf{No. of docs} & \textbf{\%}  \\
    \hline\hline
    %\midrule
USA         & 340       & 25.66  \\
China       & 264       & 19.925 \\
UK          & 122       & 9.207  \\
India       & 99        & 7.472  \\
Canada      & 61        & 4.604  \\
Italy       & 59        & 4.453  \\
Germany     & 53        & 4      \\
Australia   & 42        & 3.17   \\
France      & 37        & 2.792  \\
Netherlands & 36        & 2.717 \\
    \hline
  
   %\hline
  \end{tabular}
  
  \label{table4}
\end{table}

\subsection{Logistic Regression}
Logistic regression (LR) is a machine learning model commonly used for binary classification problems. It assumes that the data obeys a continuous probability distribution, and uses maximum likelihood estimation to estimate the parameters \cite{kleinbaum2002logistic}. LR is theoretically supported by the linear regression. The difference is that LR introduces non-linear factors through the Sigmoid function, so it can easily handle the 0/1 classification problem. It outputs the mathematical logic of the result (Eq. (\ref{equ:logistic_1})). 

\begin{equation}\label{equ:logistic_1}
\hat{p}=h_{\theta}(x)=\sigma\left(\theta^{T} \cdot x\right)
\end{equation}

The logical model (also known as logit) is a sigmoid function, denoted as , and its output is a number between 0 and 1. The definition is shown in Eq. (\ref{equ:logistic_2}).

\begin{equation}\label{equ:logistic_2}
\sigma(t)=\frac{1}{1+\exp (-t)}
\end{equation}

Once the logistic regression model estimates the probability that the sample x is classified the positive class, then the prediction y can be easily made (Eq. (\ref{equ:logistic_3}))

\begin{equation}\label{equ:logistic_3}
\hat{y}=\left\{\begin{array}{ll}0 & \hat{p} <0.5 \\ 1 & \hat{p} \geq 0.5\end{array}\right.
\end{equation}

\subsection{Decision Tree}
The decision tree (DT) is a commonly used classification method in supervised learning. DT generates a tree-like structure through the splitting of nodes and thresholds, and judges the category of input samples.The goal of DT is to predict the outcome of a sample by learning simple decision rules from data features \cite{myles2004introduction}.

\begin{figure}[ht]
\centering\includegraphics[width=0.8\linewidth]{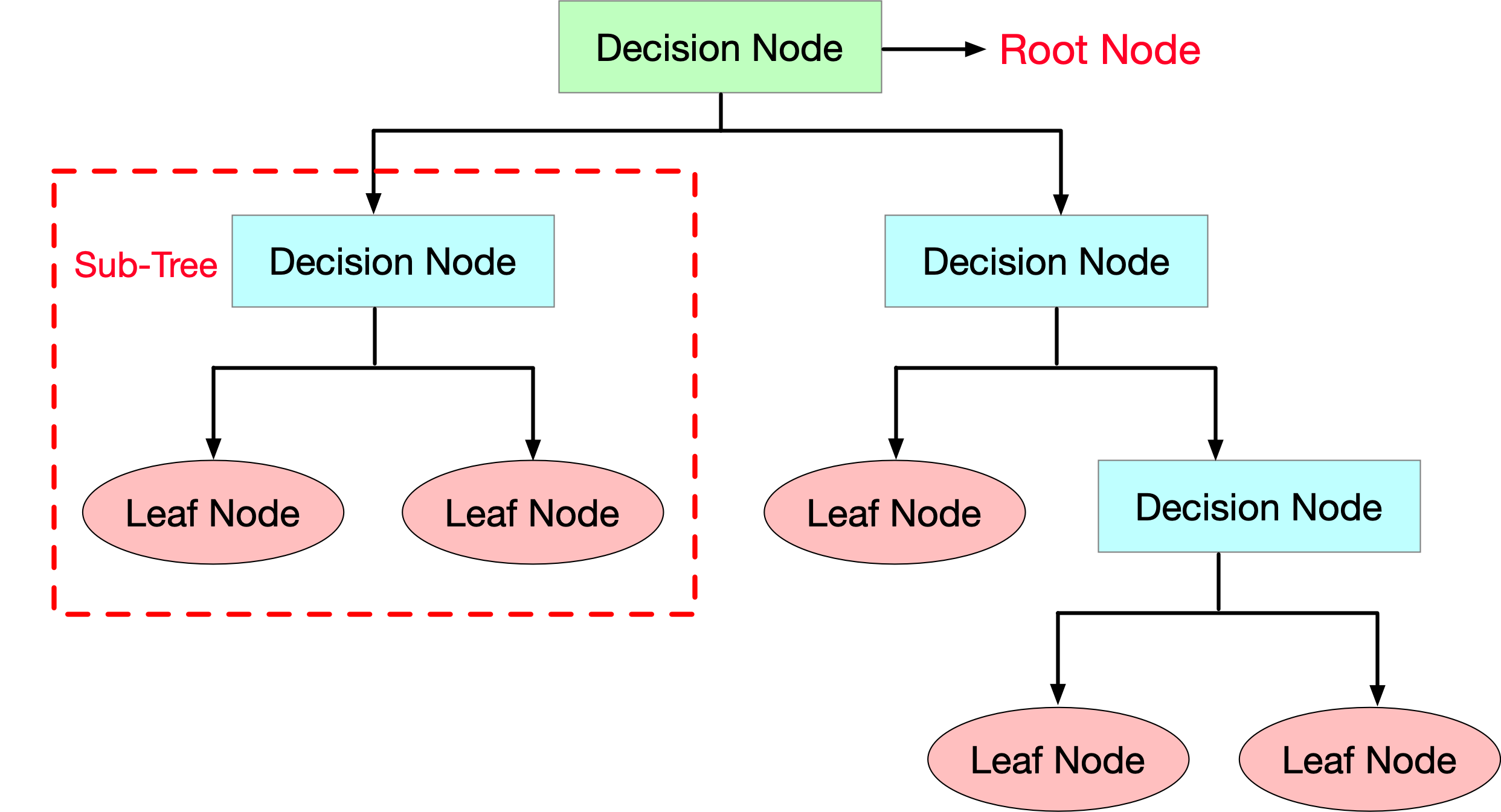}
\caption{Visualization of decision tree}
\label{decision_tree}
\end{figure}

Figure \ref{decision_tree} shows the architecture of DT. The process of constructing a complete decision tree is the process of selecting what attributes are the nodes. There are three kinds of nodes in the decision tree: root node, internal node and leaf node. Root node and internal node are the nodes that make decisions. Leaf node is the decision result. There is a parent-child relationship between nodes. Which attributes are selected as decision nodes and when to stop splitting determines the generalization performance of a DT. Controlling the depth of the DT is also a commonly used method in the modeling process.

\subsection{Gradient Boosted Decision Tree}
\begin{figure}[ht]
\centering\includegraphics[width=0.8\linewidth]{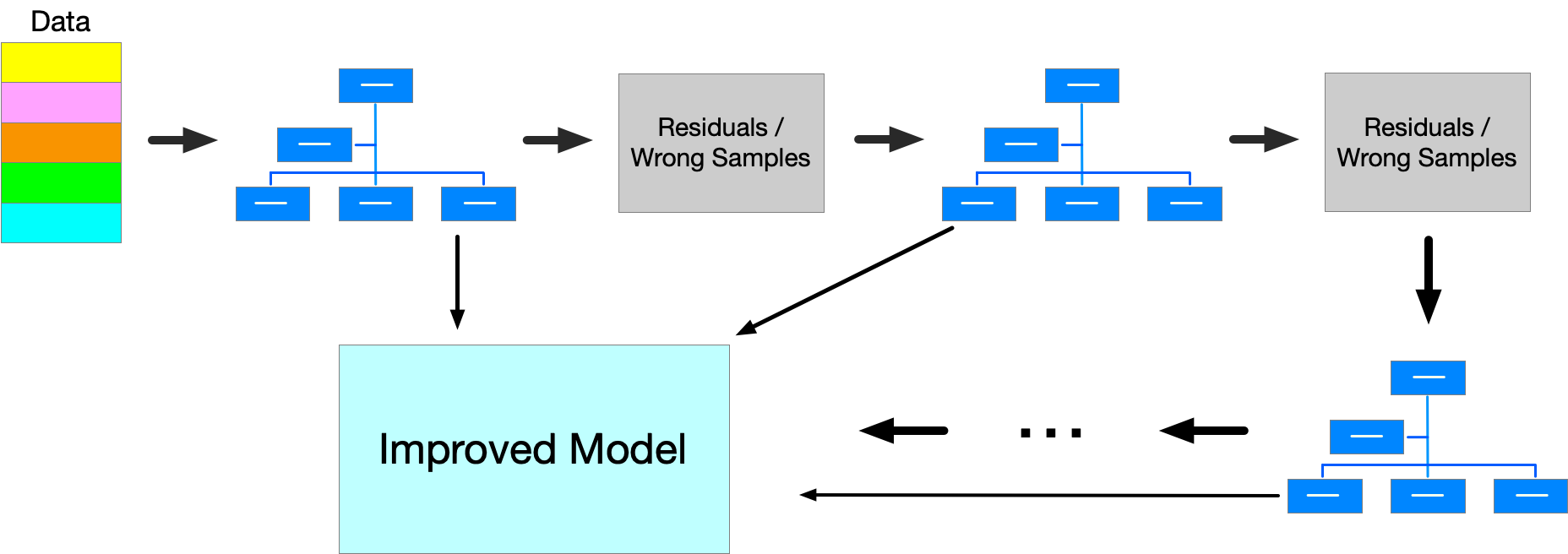}
\caption{The architecture of GBDT}
\label{gbdt}
\end{figure}

Gradient Boosted Decision Tree (GBDT) is a traditional machine learning algorithm and one of the best algorithms that fits the real distribution.It uses DT as weak learners and uses Gradient Boosting strategy for training. As shown in Figure \ref{gbdt}, it describes how GBDT works.

GBDT will conduct multiple rounds of training, and each round of training is carried out on the basis of the residual of the previous round of training. The residual here is the negative gradient value of the current model. This requires that the residual subtraction of the output of the weak classifier is meaningful during each iteration. The result of the GBDT model is a combination of a set of classification decision trees. The final output of the GBDT model is the sum of the results of a sample in each tree \cite{ke2017lightgbm}.  

\subsection{SVM}
Support Vector Machine (SVM) is a linear classification model that maximizes the interval defined in the feature space. It is commonly used in classification tasks \cite{pal2010feature}. 

\begin{figure}[ht]
\centering\includegraphics[width=0.8\linewidth]{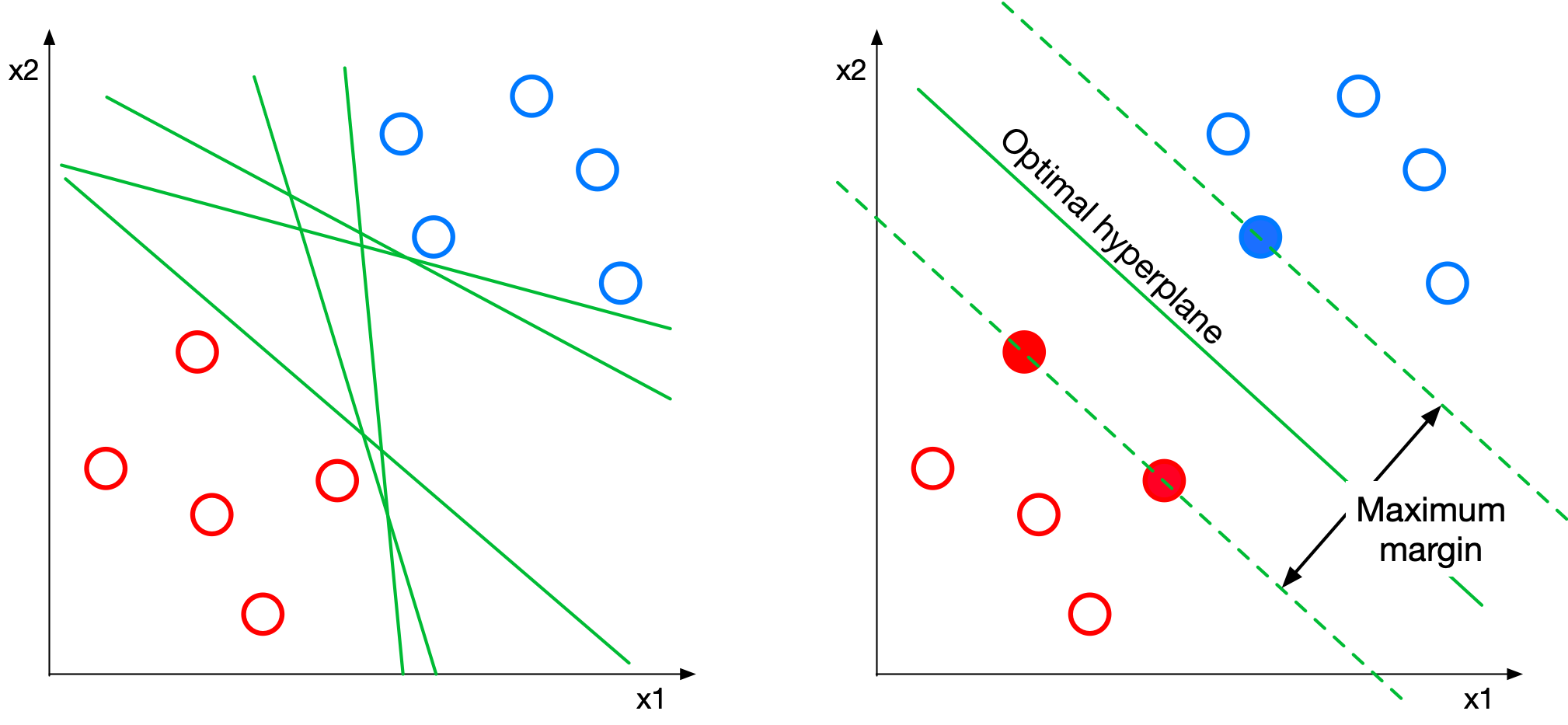}
\caption{Support Vector Machine (Left: Possible hyperplanes; Right: Maximum margin and support vectors)}
\label{svm}
\end{figure}

The goal of SVM is to find the separating hyperplane that can correctly divide the training data set and maximize the geometric interval. As shown in Figure \ref{svm}, for a linearly separable dataset, there are many possible hyperplanes, but there is one optimal hyperplane with the largest geometric interval. The data points that are at the edge of the hyperplane are support vectors. The loss function used in SVM is hinge loss, which can be defined as:

\begin{equation}\label{equ:svm}
c(x, y, f(x))=\left\{\begin{array}{lr}0, & \text { if } y * f(x) \geq 1 \\ 1-y * f(x), & \text { else }\end{array}\right.
\end{equation}
 
For $y * f(x) \geq 1$ , hinge loss is ‘0’. However, when $y * f(x)<1$ , then hinge loss increases massively. With the loss function, SVM takes partial derivatives concerning the weights to find the gradients and update the weights after. SVM will use the the regularization parameter to update the gradient when a misclassification is found.

\subsection{Multilayer Perceptron}
Multilayer perceptron (MLP) is also called Artificial Neural Network (ANN). In addition to the input and output layers, MLP can have one or more hidden layers in between. The simplest MLP only contains one hidden layer, which is a three-layer structure. 

\begin{figure}[ht]
\centering\includegraphics[width=0.7\linewidth]{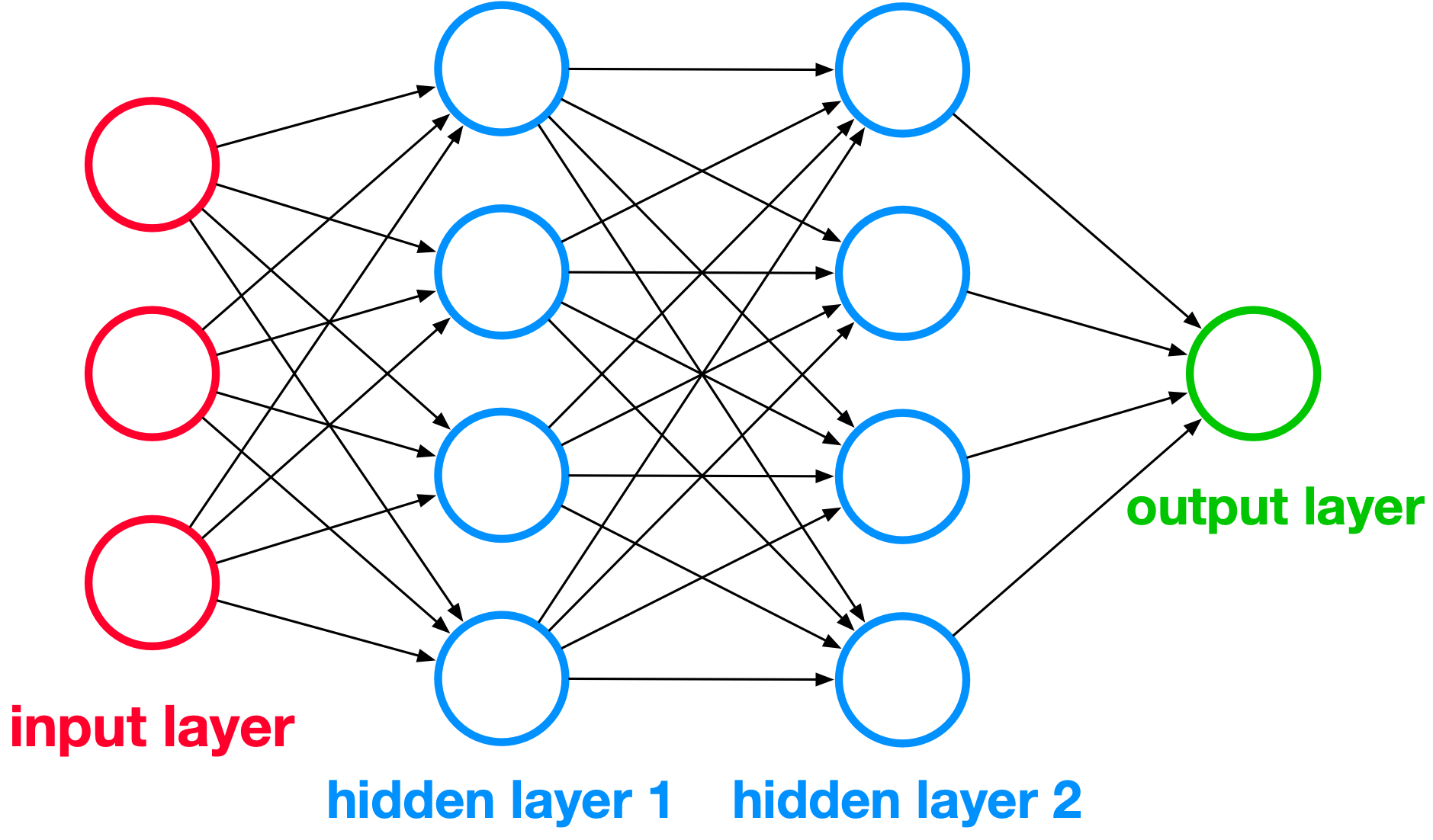}
\caption{Schematic diagram of MLP}
\label{mlp}
\end{figure}

As shown in Figure \ref{mlp}, in MLP, information is gradually transferred from the input layer to the forward layers. Except the input layer, all the neurons in the other layers use nonlinear activation function. At the beginning of training process, the instance feature vector of the training set is passed to the input layer, and then the weight of the connected node is passed to the next layer. The output of the previous layer is the input of the next layer. After the output is compared with the known label, MLP adjusts the weight accordingly (the weight usually starts with a random initialization value). This process will repeat until the model reach the maximum number of allowed iterations or an acceptable error rate.

\section{Technical Approach}
\label{S:3}

Five popular classifiers (i.e. LR, DT, GBDT, SVM, and MLP) are built and compared using their predictive accuracy on the retained data samples. 

\subsection{Dataset Description}
Research data are collected from multi-center inpatient department, including Shanghai Tenth People’s Hospital, Hainan Maternal and Children’s Medical Center, Maternity Service Center of Pengzhou Maternal \& Child Health Care Hospital, Huai’an First People’s Hospital, Nanjing Medical University. We performed a retrospective AI-based analysis of medical records of patients with MP pneumonia. Patients had been hospitalised and confirmed according to systems, signs and laboratory data of patients. Characteristics analysed included demographics (age, gender ), clinical presentation (pulmonary symptoms, clinical examination:general conditions, fever, continuous days of fever, cough, apastia or dehydration, disturbance of consciousness, respiratory rate, cyanosis, groaning, nasal flaring, three concave sign, lung infiltration area, pleural effusion, oxygen saturation, extrapulmonary complications, lung auscultation,visual examination, palpation, percussion, serum IgM and IgG of acute and convalescence, PCR test of nasopharyngeal/oropharyngeal (NP/OP) swabs; biochemistry: hepatic function(alanine amino-transferase (ALT) and aspartate amino-transferase (AST) and renal function(blood urea nitrogen (BUN), creatinine (CREA), lactate dehydrogenase (LDH), creatine kinase MB (CK-MB), leukocyte, neutrophil, lymphocyte, platelet, erythrocyte sedimentation rate(ESR) , procalcitonin (PCT), serum ferritin(SF), C-reactive protein (CRP), D-dimer, treatment with normal macrolide antibiotics for more than seven days. A CAP patient with a positive MP PCR NP/OP specimen or serum IgM titre more than 1:160 is considered to have MPP. The severity of MPP is consistent with CAP[16,17]. Hepatic and renal function (ALT,AST,BUN,CREA,CK-MB) means whether there is any damage outside the lung system when infected MP. Leukocyte, neutrophil, lymphocyte, platelet, erythrocyte sedimentation rate (ESR) , procalcitonin (PCT), serum ferritin (SF), C-reactive protein (CRP), D-dimer, treatment with normal macrolide antibiotics for more than seven days indicate the severity of the infection[18].All data including clinical symptoms, signs and biochemistry are given in Table\ref{feature}.

\begin{longtable}[c]{p{4cm} p{4cm} p{4cm} }
 \caption{Clinical variables and Significance in the diagnosis of MPP\label{long}}\\

\toprule
 \textbf{Clinical Variables}                                                      & \textbf{Reference  Range}                                                                                                               & \textbf{Clinical Significanc}\\

 \endfirsthead

 \hline
 \multicolumn{3}{ c }{Continuation of Table \ref{long}}\\
 \hline
 Clinical Variables                                                      & Reference  Range                                                                                                               & Clinical Significanc\\
 \hline
 \endhead

 \hline
 \endfoot

 \hline

 \endlastfoot
    %\multicolumn{2}{c}{Part}                   

\hline
Age                                                                     & 0.1-13year                                                                                                                    & -                                                  \\
\hline
Gender                                                                  & 0:female, 1:male                          \                                                                             & -                                                  \\
\hline
Seasons                                                                 & 0:spring, 1:summer, 2:autumn, 3:winter                                                                                  & -                                                  \\
\hline
General Condition                                                       & 0:good, 1:poor condition, 2:very bad condition                                                                                 & Judge the severity of pneumonia                    \\
\hline

Classification of Fever                                                 & 0:normal, 1:low heat, 2:moderate heat, 3:hyperpyrexia, 4:ultra-hyperpyrexia & Ultra-hyperpyrexia indicates the severity          \\
\hline
Fever With More Than  Seven Days                                                                                                                                                                                  &  0:no, 1:yes                                                                                                                    & it indicates the severity and refractory           \\
\hline
Cough Nature                           
                                         & 0:without cough, 1:dry cough,  2:phlegmy cough                                                                                                 & A symptom                                           \\
\hline
Apastia or Dehydration                                                  & 0:no, 1:yes                                                                                                                    & Judge the severity                                 \\
\hline
Consciousness Disturbance                                               & 0:no, 1:yes                                                                                                                    & Judge the severity                                 \\
\hline
Respiratory Rate                                                        & 0:normal, 1:breathing faster,                                                                                                                
                                                                        2:breathing noticeably faster                                                                                                  &    Judge the severity                                                  \\
 \hline
Cyanosis                                                                & 0:no, 1:yes                                                                                                                    & Judge the severity                                 \\
\hline
Groaning                                                                & 0:no, 1:yes                                                                                                                    & Judge the severity                                 \\
\hline
Nasal Flaring                                                           & 0:no, 1:yes                                                                                                                    & Judge the severity                                 \\
\hline
Three Concave Sign                                                      & 0:no, 1:yes                                                                                                                    & Judge the severity                                 \\
\hline
Lung Infiltration Area                                                  & 0:no, 1:mild, 2:severe                                                                                                         & Judge the severity                                 \\
\hline
Pleural Effusion                                                        & 0:no, 1:yes                                                                                                                    & Judge the severity                                 \\
\hline
Pulse Blood Oxygen Saturation                                           & 0:normal, 1:abnormal                                                                                                           & Judge the severity                                 \\
\hline
Extropulmonary   Complications                                                           & 0:no, 1:yes                                                                                                                    & Judge the severity                                 \\
                
\hline
Rhonchus or Wheeze                                                      & 0:no, 1:left, 2:right, 3:left and right                                                                                        & The sign of the disease                            \\
\hline
Moist Crackles                                                          & 0:no, 1:left, 2:right, 3:left and right                                                                                        & The sign of the disease                            \\
\hline
Decreased Breath Sounds                                                 & 0:no;1:yes                                                                                                                     & The sign of the disease                            \\
\hline
Reduced Breathing Movement                                              & 0:no;1:yes                                                                                                                     & The sign of the disease                            \\
\hline
Enhancement Tactile Fremitus                                            & 0:no;1:yes                                                                                                                     & The sign of the disease                            \\
\hline
Lung Percussion                                                         & 0:normal;1:dullness, 2:flatness                                                                                                & The sign of the disease                            \\
\hline
Pleural Friction Rub                                                    & 0:no, 1:yes                                                                                                                    & The sign of the disease                            \\
\hline
IgM(acute stage)                                                        & -1:not check, 0:negative;1:positive                                                                                            & Positive indicates the infection of MP             \\
\hline
IgG(acute stage)                                                        & 0:negative, 1:positive                                                                                                         & Positive indicates infected in the past            \\
\hline
IgM (Convalescence)                                                     & -1:not check, 0:negative;1:positive                                                                                            & -                                                  \\
\hline
IgG (Convalescence)                                                     & 0:negative, 1:positive                                                                                                         & Positive indicates antibody production             \\
\hline
Nucleic Acid PCR(NP/OP)                                                 & -1:not check;0:negative;1:positive                                                                                             & Positive indicates the infection of MP             \\
\hline
ALT( Alanine Amino- transferase)                                        & 0-40u/L                                                                                                                        & Indicates whether hepatic function damage          \\
\hline
AST(Aspartate Amino- transferase)                                       & 0-40U/L                                                                                                                        & Indicates whether hepatic function damage          \\
\hline
BUN(Blood Urea Nitrogen)                                                & 1.8-6.5 mmol/L                                                                                                                 & Indicates whether kidney function damage           \\
\hline
CREA(Creatinine)                                                        & 24.9-69.7mmol/L                                                                                                                & Indicates whether kidney function damage           \\
\hline
LDH(Lactate Dehydrogenase)                                              & 100-300U/L                                                                                                                     & Indicates whether myocardial injury and refractory \\
\hline
CK-MB(creatine kinase MB)(mass)                                         & \textless{}0.6ng/ml                                                                                                            & Indicates whether myocardial damage                \\
\hline
CK-MB(activity)                                                         & 0-24U/L                                                                                                                        & Indicates whether myocardial injury                \\
\hline
Leukocyte                                                               & (4-10)*109                                                                                                                     & Indicates infection                                \\
\hline
Neutrophil\%                                                            & 43\%-76\%                                                                                                                      & Indicates infection                                \\
\hline
Lymphocyte\%                                                            & 17\%-48\%                                                                                                                      & Indicates infection                                \\
\hline
Platelet                                                                & (100-300)*109                                                                                                                  & Indicates infection                                \\
\hline
CRP(C Reactive Protein)                                                 & \textless{}10mg/L                                                                                                              & Indicates infection                                \\
\hline
ESR(Erythrocyte Sedimentation Rate)                                     & male:0-15mm/h;female:0-20mm/h                                                                                                  & Indicates infection                                \\
\hline
PCT(Procalcitonin)                                                      & \textless{}0.5ng/ml                                                                                                            & Indicates infection                                \\
\hline
SF(Serum Ferritin)                                                      & female:12-150ug/L;male:15-200ug/L                                                                                              & Indicates infection                                \\
\hline
D-Dimer                                                                 & \textless{}0.5mg/L                                                                                                             & Indicates whether hypercoagulability               \\
\hline
After seven days or more of normal treatment with macrolide antibiotics & 0:no;1:persistent fever;2:aggravated clinical signs;3:extrapulmonary complications;4:progression in imaging;5:extension course & It indicates the severity and refractory  
\label{feature}
\end{longtable}

\subsection{Prediction Models}
\begin{figure}[H]
\centering\includegraphics[width=0.8\linewidth]{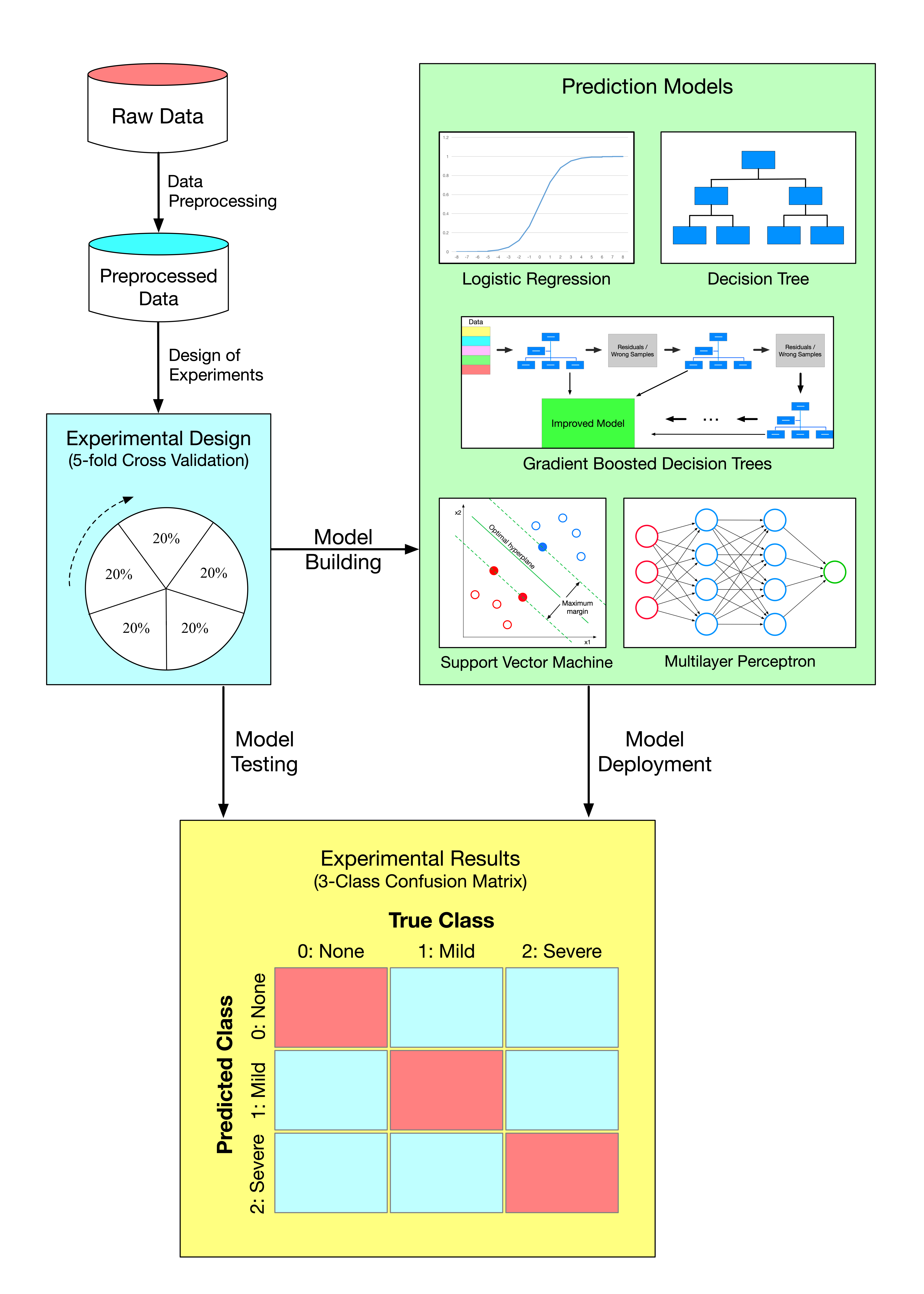}
\caption{Workflow of the technical approach}
\label{workflow}
\end{figure}

In this study, the proposed method consists of four stages. In the first stage, several feature engineering operations, such as feature selection and feature cleaning, are conducted to obtain an improved dataset. In the second stage, a 5-fold cross-validation approach is utilized to estimate the performance of the prediction models. In cross validation, 5 is one of the optimal number for folds. It can effectively reduce the deviation and variance generated in the verification process, and can also shorten the test time. The original dataset is split into five independent subsets in our approach, and the subsets are directly mutually exclusive. Each fold is used separately to test the performance of the model, and finally five performance estimates are obtained. In the third stage, we test the machine learning models that are selected in our approach. Since we use a 5-fold cross-validation, each model will perform five independent experiments. Finally, the classification performance of the five machine learning models are compared. The pictorial depiction of our proposed method is shown in Figure \ref{workflow}. 

In this approach, five classic classification methods (i.e., LR, DT, GBDT, SVM and MLP) are built and compared to each other using their predictive accuracy on the retained samples. An LR model is built to predict the odds of the mycoplasma pneumonia occurrence, instead of predicting a point estimate of the disease itself. A DT model is constructed by asking a series of questions with respect to a record of the pneumonia dataset we have got. A GBDT model conducts an optimized loss function and use decision-tree as a weak learner. Each decision tree makes predictions. Then an additive model is built to add them to minimize the loss function. SVM uses hinge loss function to achieve the maximum margin hyperplane to distinguish data points belong to different classes. MLP uses forward propagation to continuously update model‘s weights until the best prediction performance is obtained.

\subsection{Evaluation Metrics}
In machine learning algorithms, confusion matrix is a method to use a specific matrix to present the visualization of algorithm's performance. Each column represents the predicted value, and each row represents the actual category \cite{godbole2004discriminative}. In this paper, our classification problem is a 3-class classification problem. The prediction result for each sample will be: 0, 1 or 2. 0 means the patient is not infected, 1 means the patient is mildly infected, and 2 means the patient is severely infected. 

We conduct the confusion matrix for 3-class classification in our study. Unlike positive or negative classes in binary classification, we aim to find the metrics of the confusion matrix for each individual class. For example, if we take class 1 (mildly infected), the class 0 and 2 will be combined as an opposite class. With the confusion matrix generated, we can calculate the performance measure for class 1. Similarly, we can calculate the measures for the other two classes. 

In our study, we present various performance measures to evaluate and compare the five models for mycoplasma pneumonia diagnosis. Accuracy gives the fraction of the total samples that were correctly classified by the classifier. Precision reflects the model's ability to distinguish negative samples, and recall reflects the classification model's ability to recognize positive samples. F1-score is a combination of precision and recall. It indicates the robustness of the classification model. From a perspective of the confusion matrix, the formula for calculating accuracy, precision, recall, and f1-score are defined as follows, where TP is Ture Positive, TN is True Negative, FP is False Positive , and FN is False Negative: 

\begin{equation}\label{equ:acc}
\text { Accuracy }=\frac{T P+T N}{T P+T N+F P+F N}
\end{equation}
\begin{equation}\label{equ:precision}
\text { Precision }=\frac{T P}{T P+F P}
\end{equation}
\begin{equation}\label{equ:recall}
\text { Recall }=\frac{T P}{T P+F N}
\end{equation}
\begin{equation}\label{equ:fone}
\mathrm{F} 1 \text { score }=\frac{2 * \text { Precision } * \text { Recall }}{\text {Precision }+\text { Recall }}
\end{equation}

\section{Results and Discussion}
\label{S:4}
The training and test experiments are conducted on our own dataset, which is composed of 960 records.The experimental results show that GBDT has the best performance among the five methods with an overall accuracy rate of 0.937.

\subsection{Prediction Performance}
\begin{figure}[ht]
\centering\includegraphics[width=0.8\linewidth]{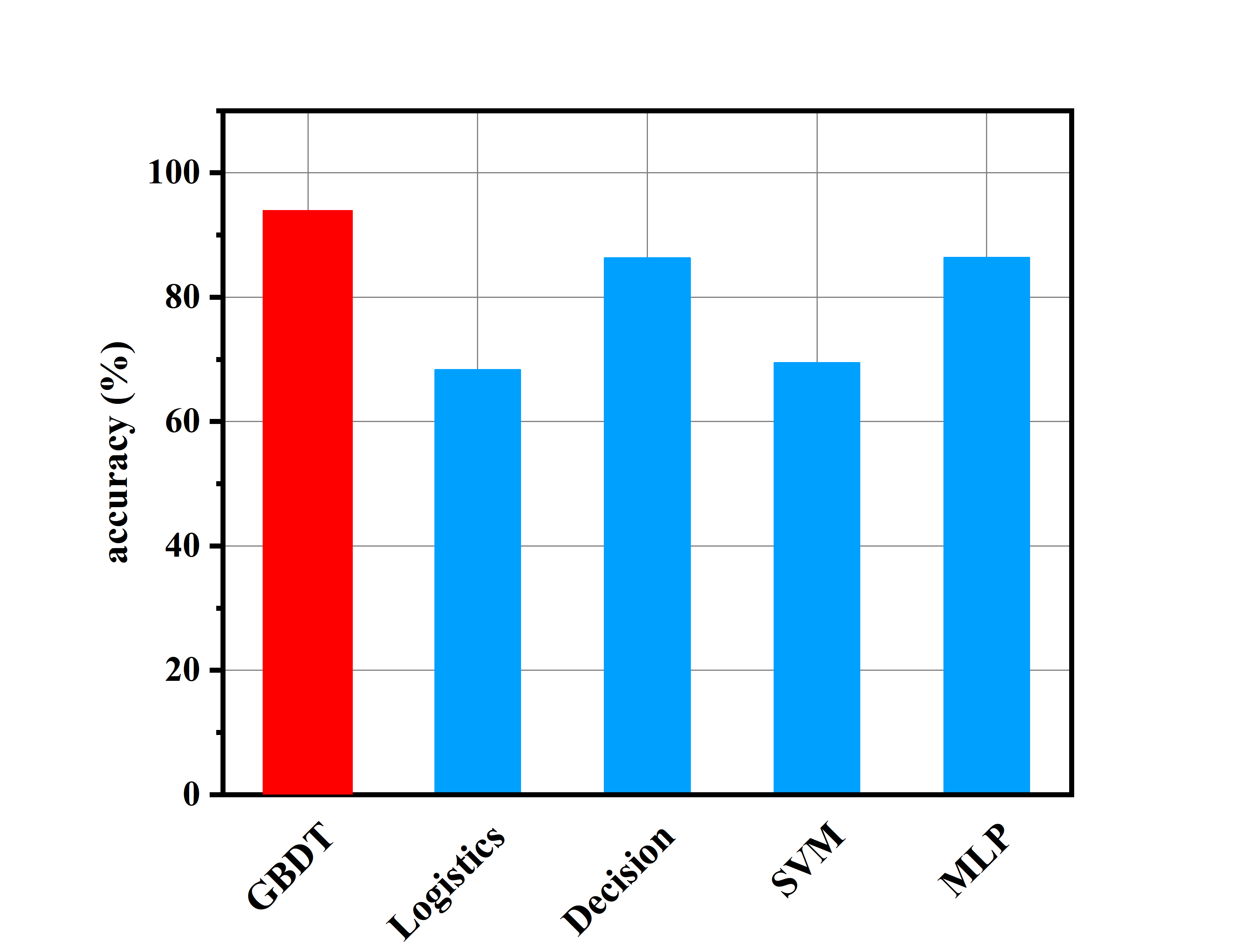}
\caption{Comparison of the accuracy results obtained by machine learning models}
\label{accuracy}
\end{figure}

We use the original dataset which was composed of 960 records in the experiments. As shown in Figure \ref{accuracy}, based on the 5-fold cross-validation, the GBDT produces the best results with an overall accuracy rate of 0.937, and the decision tree comes out as the second best with an overall accuracy of 0.884, followed by MLP and logistic regression with overall accuracy rates of 0.863 and 0.695 respectively. SVM achieves an overall accuracy of 0.653 which is not as good as other models.

The other classification results are given in Table \ref{0}, \ref{1} and \ref{2}. The best result is obtained by GBDT. In practice, we pay more attention to the predictive performance of mild and severe infections. The precision and recall of class “1:mild infected” reaches 0.946 and 0.946 in GBDT, while the precision and recall of class “2:severe infected” reaches 1 and 0.895. The F1-score of the three classes (i.e., 0:none, 1:mild, 2:severe ) in GBDT are 0.905,0.946 and 0.944 .
\begin{table}[H]
  \centering
  \caption{No infected'0'}
  \begin{tabular}{l l l l }
   \hline
    \toprule
    %\multicolumn{2}{c}{Part}                   
    \cmidrule(r){1-2}
    \textbf{Model}  & \textbf{Precision} & \textbf{Recall} & \textbf{F1 score}  \\
    \hline\hline
    %\midrule
    GBDT & 0.863 & 0.951 & 0.905  \\
    Logistics  & 0.932 & 0.178 & 0.611  \\
    
    Decision tree & 0.944 & 0.851 & 0.895   \\
    
    SVM  & 0.875 & 0.35 & 0.5   \\
    
    MLP  & 0.81 & 0.952 & 0.869  \\
    \bottomrule
   \hline
  \end{tabular}
  
  \label{0}
\end{table}

\begin{table}[H]
  \centering
  \caption{Mild infected'1'}
  \begin{tabular}{l l l l }
   \hline
    \toprule
    %\multicolumn{2}{c}{Part}                   
    \cmidrule(r){1-2}
   \textbf{Model}  & \textbf{Precision} & \textbf{Recall} & \textbf{F1 score}  \\
    \hline\hline
    %\midrule
    GBDT & 0.946 & 0.946 & 0.946  \\
    Logistics  & 0.704 & 0.892 & 0.787  \\
    
    Decision tree & 0.961 & 0.892 & 0.925   \\
    
    SVM  & 0.638 & 0.946 & 0.962   \\
    
    MLP  & 0.962 & 0.716 & 0.821  \\
    \bottomrule
   \hline
  \end{tabular}
  
  \label{1}
\end{table}

\begin{table}[H]
  \centering
  \caption{Severe infected'2'}
  \begin{tabular}{l l l l }
   \hline
    \toprule
    %\multicolumn{2}{c}{Part}                   
    \cmidrule(r){1-2}
   \textbf{Model} & \textbf{Precision} & \textbf{Recall} & \textbf{F1 score} \\
    \hline\hline
    %\midrule
    GBDT & 1 & 0.895 & 0.944  \\
    Logistics & 0.625 & 0.263 & 0.370 \\
    
    Decision tree & 0.680 & 0.895 & 0.773 \\
    
    SVM & 0.510 & 0.105 & 0.174 \\
    
    MLP & 0.644 & 0.974 & 0.776 \\
    \bottomrule
   \hline
  \end{tabular}
  \label{2}
\end{table}

Since the decision rules of our machine learning models are hardly accessible to humans and cannot easily be used to gain insights about the mycoplasmal pneumonia diagnosis in children. Therefore, we display the Feature Importance in our study, which by retrospective analysis of arbitrary learning machines achieve both excellent predictive performance and superior interpretation. 

In contrast to standard raw feature weighting, the feature importance takes the underlying correlation structure of the features into account. Thereby, it is able to discover the most relevant features, even if their appearance in the training data is entirely prevented by noise. 

We list the top 10 features ranked by feature importance scores in GBDT. As shown in Table \ref{importance}, the most important feature is the “Pulmonary infiltrates range” with the score of 0.5925, follow by “Cough” (0.0953) and “Pleural effusion” (0.0492). “Pulmonary signs” is also an important feature that cannot be ignored. 

The chest radiographic examination is an essential part of the diagnosis of pneumonia including MPP. Furthermore, chest radiographs play an important role in assessing a patient’s current condition and prognosis, as well as in determining the treatment plan, which is consistent with our study result. In the present day, Pulmonary infiltrates range is the most important clinical variable, it demonstrates that it is an important role in evaluating the severity of MPP. Cough is a common symptom of respiratory disease including MPP. So It is a sensitive but poorly specific indicator of MPP \citet{wang2012clinical}.In our study, it also shows that it plays an important role in the diagnosis of MPP. Pleural effusion is an important feature in helping clinicians to distinguish the mild and severity of MPP. Dry rale and wheezing are also a vital characteristic of MPP, it perhaps confirms the previous study result indirectly that a small part of children suffered from recurrent wheezing after MPP and children increased the risk of wheezing experience when they had MP infection \cite{rhim2019epidemiological}. Wet rale is a common clinical sign in pneumonia,and is also a diagnostic indicator of MPP. Other characters, such as WBC, CRP, renal function (CREA and BUN) and CK-MB, are also indicators in estimating the severity and prognosis of MPP.

\begin{table}[H]
  \centering
  \caption{The top 10 features ranked by feature importance scores in GBDT}
  \begin{tabular}{l l l }
   \hline
    \toprule
    %\multicolumn{2}{c}{Part}                   
    \cmidrule(r){1-3}
   \textbf{Rank} & \textbf{Name of Feature}                                        & \textbf{Feature Importance} \\
\hline\hline
1    & Pulmonary infiltrates range                            & 0.5925,            \\
2    & Cough                                                  & 0.0953             \\
3    & Pleural effusion                                       & 0.0492             \\
4    & Pulmonary sign, whether there are dry rales, wheezing & 0.0453             \\
5    & Pulmonary sign, whether there is wet rale           & 0.0397             \\
6    & White blood cells (WBC)                                & 0.0195             \\
7    & Renal function (CREA, Umol/L)                          & 0.0154             \\
8    & CK-MB, U/L (active)                                    & 0.0152             \\
9    & C-reactive protein (CRP) mg/L                          & 0.0131             \\
10   & Renal function (BUN, mmol/L)                           & 0.0104 \\
    \bottomrule
   \hline
  \end{tabular}
  
  \label{importance}
\end{table}

\section{Conclusions}
\label{S:5}
In this work, we utilize LR, DT, GBDT, SVM and MLP models to rapidly predict MPP diagnosis in children patients. We collect and organize the raw MPP dataset from five different center inpatient departments within the past two years. We employ the preprocessing procedure to the original dataset to ensure the best prediction effectiveness of the models. After the classification tasks, we conduct a three-class confusion matrix in the evaluation of our prediction experiments. GBDT outperforms other machine learning models in terms of all the three classes (i.e., no infected, mild infected, severe infected). It achieves the highest accuracy of 93.7\%. 

Finally, through the feature importance analysis, we list the most importance features in our study. “pulmonary infiltrates range” is the most important feature of GBDT with the score of 0.5925, follow by “cough” (0.0953) and “pleural effusion” (0.0492). 

In our future work, we will implement these machine learning methods on a larger dataset with more children MPP cases. Data mining and deeper feature correlation analysis will be taken into account, in order to obtain better prediction performance.

\section*{Compliance with Ethical Standards}
\label{S:6}
This article does not contain any studies with human participants or animals performed by any of the authors.



%% References


%% New version of the num-names style
\bibliographystyle{elsarticle-num-names}

%\bibliography{references.bib}

\begin{thebibliography}{26}
\expandafter\ifx\csname natexlab\endcsname\relax\def\natexlab#1{#1}\fi
\providecommand{\url}[1]{\texttt{#1}}
\providecommand{\href}[2]{#2}
\providecommand{\path}[1]{#1}
\providecommand{\DOIprefix}{doi:}
\providecommand{\ArXivprefix}{arXiv:}
\providecommand{\URLprefix}{URL: }
\providecommand{\Pubmedprefix}{pmid:}
\providecommand{\doi}[1]{\href{http://dx.doi.org/#1}{\path{#1}}}
\providecommand{\Pubmed}[1]{\href{pmid:#1}{\path{#1}}}
\providecommand{\bibinfo}[2]{#2}
\ifx\xfnm\relax \def\xfnm[#1]{\unskip,\space#1}\fi
%Type = Article
\bibitem[{Biondi et~al.(2014)Biondi, McCulloh, Alverson, Klein, Dixon, and
  Ralston}]{biondi2014treatment}
\bibinfo{author}{E.~Biondi}, \bibinfo{author}{R.~McCulloh},
  \bibinfo{author}{B.~Alverson}, \bibinfo{author}{A.~Klein},
  \bibinfo{author}{A.~Dixon}, \bibinfo{author}{S.~Ralston},
\newblock \bibinfo{title}{Treatment of mycoplasma pneumonia: a systematic
  review},
\newblock \bibinfo{journal}{Pediatrics} \bibinfo{volume}{133}
  (\bibinfo{year}{2014}) \bibinfo{pages}{1081--1090}.
%Type = Article
\bibitem[{Bradley et~al.(2011)Bradley, Byington, Shah, Alverson, Carter,
  Harrison, Kaplan, Mace, McCracken~Jr, Moore et~al.}]{bradley2011management}
\bibinfo{author}{J.~S. Bradley}, \bibinfo{author}{C.~L. Byington},
  \bibinfo{author}{S.~S. Shah}, \bibinfo{author}{B.~Alverson},
  \bibinfo{author}{E.~R. Carter}, \bibinfo{author}{C.~Harrison},
  \bibinfo{author}{S.~L. Kaplan}, \bibinfo{author}{S.~E. Mace},
  \bibinfo{author}{G.~H. McCracken~Jr}, \bibinfo{author}{M.~R. Moore}, et~al.,
\newblock \bibinfo{title}{The management of community-acquired pneumonia in
  infants and children older than 3 months of age: clinical practice guidelines
  by the pediatric infectious diseases society and the infectious diseases
  society of america},
\newblock \bibinfo{journal}{Clinical infectious diseases} \bibinfo{volume}{53}
  (\bibinfo{year}{2011}) \bibinfo{pages}{e25--e76}.
%Type = Article
\bibitem[{Korppi et~al.(2004)Korppi, Heiskanen-Kosma, and
  Kleemola}]{korppi2004incidence}
\bibinfo{author}{M.~Korppi}, \bibinfo{author}{T.~Heiskanen-Kosma},
  \bibinfo{author}{M.~Kleemola},
\newblock \bibinfo{title}{Incidence of community-acquired pneumonia in children
  caused by mycoplasma pneumoniae: serological results of a prospective,
  population-based study in primary health care},
\newblock \bibinfo{journal}{Respirology} \bibinfo{volume}{9}
  (\bibinfo{year}{2004}) \bibinfo{pages}{109--114}.
%Type = Article
\bibitem[{AtkinsonTP(2008)}]{atkinsontp2008waiteskb}
\bibinfo{author}{B.~AtkinsonTP},
\newblock \bibinfo{title}{Waiteskb. epidemiology, clinical manifestations,
  pathogenesis and laboratory detection of mycoplasma pneumoniae infections},
\newblock \bibinfo{journal}{FEMS Microbiol Rev} \bibinfo{volume}{32}
  (\bibinfo{year}{2008}) \bibinfo{pages}{956--73}.
%Type = Article
\bibitem[{Lind et~al.(1997)Lind, Benzon, Jensen, and
  Clyde}]{lind1997seroepidemiological}
\bibinfo{author}{K.~Lind}, \bibinfo{author}{M.~Benzon},
  \bibinfo{author}{S.~Jensen}, \bibinfo{author}{W.~Clyde},
\newblock \bibinfo{title}{A seroepidemiological study of mycoplasma pneumoniae
  infections in denmark over the 50-year period 1946--1995},
\newblock \bibinfo{journal}{European journal of epidemiology}
  \bibinfo{volume}{13} (\bibinfo{year}{1997}) \bibinfo{pages}{581--586}.
%Type = Article
\bibitem[{Olson et~al.(2015)Olson, Watkins, Demirjian, Lin, Robinson, Pretty,
  Benitez, Winchell, Diaz, Miller et~al.}]{olson2015outbreak}
\bibinfo{author}{D.~Olson}, \bibinfo{author}{L.~K.~F. Watkins},
  \bibinfo{author}{A.~Demirjian}, \bibinfo{author}{X.~Lin},
  \bibinfo{author}{C.~C. Robinson}, \bibinfo{author}{K.~Pretty},
  \bibinfo{author}{A.~J. Benitez}, \bibinfo{author}{J.~M. Winchell},
  \bibinfo{author}{M.~H. Diaz}, \bibinfo{author}{L.~A. Miller}, et~al.,
\newblock \bibinfo{title}{Outbreak of mycoplasma pneumoniae--associated
  stevens-johnson syndrome},
\newblock \bibinfo{journal}{Pediatrics} \bibinfo{volume}{136}
  (\bibinfo{year}{2015}) \bibinfo{pages}{e386--e394}.
%Type = Article
\bibitem[{Tsiodras et~al.(2005)Tsiodras, Kelesidis, Kelesidis, Stamboulis, and
  Giamarellou}]{tsiodras2005central}
\bibinfo{author}{S.~Tsiodras}, \bibinfo{author}{I.~Kelesidis},
  \bibinfo{author}{T.~Kelesidis}, \bibinfo{author}{E.~Stamboulis},
  \bibinfo{author}{H.~Giamarellou},
\newblock \bibinfo{title}{Central nervous system manifestations of mycoplasma
  pneumoniae infections},
\newblock \bibinfo{journal}{Journal of Infection} \bibinfo{volume}{51}
  (\bibinfo{year}{2005}) \bibinfo{pages}{343--354}.
%Type = Article
\bibitem[{Diaz et~al.(2015)Diaz, Benitez, and
  Winchell}]{diaz2015investigations}
\bibinfo{author}{M.~H. Diaz}, \bibinfo{author}{A.~J. Benitez},
  \bibinfo{author}{J.~M. Winchell},
\newblock \bibinfo{title}{Investigations of mycoplasma pneumoniae infections in
  the united states: trends in molecular typing and macrolide resistance from
  2006 to 2013},
\newblock \bibinfo{journal}{Journal of clinical microbiology}
  \bibinfo{volume}{53} (\bibinfo{year}{2015}) \bibinfo{pages}{124--130}.
%Type = Article
\bibitem[{Dumke et~al.(2010)Dumke, von Baum, L{\"u}ck, and
  Jacobs}]{dumke2010occurrence}
\bibinfo{author}{R.~Dumke}, \bibinfo{author}{H.~von Baum},
  \bibinfo{author}{P.~C. L{\"u}ck}, \bibinfo{author}{E.~Jacobs},
\newblock \bibinfo{title}{Occurrence of macrolide-resistant mycoplasma
  pneumoniae strains in germany},
\newblock \bibinfo{journal}{Clinical Microbiology and Infection}
  \bibinfo{volume}{16} (\bibinfo{year}{2010}) \bibinfo{pages}{613--616}.
%Type = Article
\bibitem[{Pereyre et~al.(2013)Pereyre, Touati, Petitjean-Lecherbonnier,
  Charron, Vabret, and B{\'e}b{\'e}ar}]{pereyre2013increased}
\bibinfo{author}{S.~Pereyre}, \bibinfo{author}{A.~Touati},
  \bibinfo{author}{J.~Petitjean-Lecherbonnier}, \bibinfo{author}{A.~Charron},
  \bibinfo{author}{A.~Vabret}, \bibinfo{author}{C.~B{\'e}b{\'e}ar},
\newblock \bibinfo{title}{The increased incidence of m ycoplasma pneumoniae in
  f rance in 2011 was polyclonal, mainly involving m. pneumoniae type 1
  strains},
\newblock \bibinfo{journal}{Clinical Microbiology and Infection}
  \bibinfo{volume}{19} (\bibinfo{year}{2013}) \bibinfo{pages}{E212--E217}.
%Type = Article
\bibitem[{Ho et~al.(2015)Ho, Law, Chan, Wong, To, Chiu, Cheng, and
  Yam}]{ho2015emergence}
\bibinfo{author}{P.-L. Ho}, \bibinfo{author}{P.~Y. Law}, \bibinfo{author}{B.~W.
  Chan}, \bibinfo{author}{C.-W. Wong}, \bibinfo{author}{K.~K. To},
  \bibinfo{author}{S.~S. Chiu}, \bibinfo{author}{V.~C. Cheng},
  \bibinfo{author}{W.-C. Yam},
\newblock \bibinfo{title}{Emergence of macrolide-resistant mycoplasma
  pneumoniae in hong kong is linked to increasing macrolide resistance in
  multilocus variable-number tandem-repeat analysis type 4-5-7-2},
\newblock \bibinfo{journal}{Journal of clinical microbiology}
  \bibinfo{volume}{53} (\bibinfo{year}{2015}) \bibinfo{pages}{3560--3564}.
%Type = Article
\bibitem[{Peirano et~al.(2014)Peirano, van~der Bij, Freeman, Poirel, Nordmann,
  Costello, Tchesnokova, and Pitout}]{peirano2014characteristics}
\bibinfo{author}{G.~Peirano}, \bibinfo{author}{A.~K. van~der Bij},
  \bibinfo{author}{J.~L. Freeman}, \bibinfo{author}{L.~Poirel},
  \bibinfo{author}{P.~Nordmann}, \bibinfo{author}{M.~Costello},
  \bibinfo{author}{V.~L. Tchesnokova}, \bibinfo{author}{J.~D. Pitout},
\newblock \bibinfo{title}{Characteristics of escherichia coli sequence type 131
  isolates that produce extended-spectrum $\beta$-lactamases: global
  distribution of the h30-rx sublineage},
\newblock \bibinfo{journal}{Antimicrobial agents and chemotherapy}
  \bibinfo{volume}{58} (\bibinfo{year}{2014}) \bibinfo{pages}{3762--3767}.
%Type = Article
\bibitem[{Tanaka et~al.(2017)Tanaka, Oishi, Miyata, Wakabayashi, Kono, Ono,
  Kato, Fukuda, Saito, Kondo et~al.}]{tanaka2017macrolide}
\bibinfo{author}{T.~Tanaka}, \bibinfo{author}{T.~Oishi},
  \bibinfo{author}{I.~Miyata}, \bibinfo{author}{S.~Wakabayashi},
  \bibinfo{author}{M.~Kono}, \bibinfo{author}{S.~Ono},
  \bibinfo{author}{A.~Kato}, \bibinfo{author}{Y.~Fukuda},
  \bibinfo{author}{A.~Saito}, \bibinfo{author}{E.~Kondo}, et~al.,
\newblock \bibinfo{title}{Macrolide-resistant mycoplasma pneumoniae infection,
  japan, 2008--2015},
\newblock \bibinfo{journal}{Emerging infectious diseases} \bibinfo{volume}{23}
  (\bibinfo{year}{2017}) \bibinfo{pages}{1703}.
%Type = Article
\bibitem[{Perer et~al.(2015)Perer, Wang, and Hu}]{perer2015mining}
\bibinfo{author}{A.~Perer}, \bibinfo{author}{F.~Wang}, \bibinfo{author}{J.~Hu},
\newblock \bibinfo{title}{Mining and exploring care pathways from electronic
  medical records with visual analytics},
\newblock \bibinfo{journal}{Journal of biomedical informatics}
  \bibinfo{volume}{56} (\bibinfo{year}{2015}) \bibinfo{pages}{369--378}.
%Type = Article
\bibitem[{Zech et~al.(2018)Zech, Badgeley, Liu, Costa, Titano, and
  Oermann}]{zech2018variable}
\bibinfo{author}{J.~R. Zech}, \bibinfo{author}{M.~A. Badgeley},
  \bibinfo{author}{M.~Liu}, \bibinfo{author}{A.~B. Costa},
  \bibinfo{author}{J.~J. Titano}, \bibinfo{author}{E.~K. Oermann},
\newblock \bibinfo{title}{Variable generalization performance of a deep
  learning model to detect pneumonia in chest radiographs: a cross-sectional
  study},
\newblock \bibinfo{journal}{PLoS medicine} \bibinfo{volume}{15}
  (\bibinfo{year}{2018}) \bibinfo{pages}{e1002683}.
%Type = Article
\bibitem[{Ge et~al.(2019)Ge, Wang, Wang, Wu, Peng, Wang, Xu, Xiong, Zhang, and
  Yi}]{ge2019predicting}
\bibinfo{author}{Y.~Ge}, \bibinfo{author}{Q.~Wang}, \bibinfo{author}{L.~Wang},
  \bibinfo{author}{H.~Wu}, \bibinfo{author}{C.~Peng},
  \bibinfo{author}{J.~Wang}, \bibinfo{author}{Y.~Xu},
  \bibinfo{author}{G.~Xiong}, \bibinfo{author}{Y.~Zhang},
  \bibinfo{author}{Y.~Yi},
\newblock \bibinfo{title}{Predicting post-stroke pneumonia using deep neural
  network approaches},
\newblock \bibinfo{journal}{International Journal of Medical Informatics}
  \bibinfo{volume}{132} (\bibinfo{year}{2019}) \bibinfo{pages}{103986}.
%Type = Article
\bibitem[{Stephen et~al.(2019)Stephen, Sain, Maduh, and
  Jeong}]{stephen2019efficient}
\bibinfo{author}{O.~Stephen}, \bibinfo{author}{M.~Sain}, \bibinfo{author}{U.~J.
  Maduh}, \bibinfo{author}{D.-U. Jeong},
\newblock \bibinfo{title}{An efficient deep learning approach to pneumonia
  classification in healthcare},
\newblock \bibinfo{journal}{Journal of healthcare engineering}
  \bibinfo{volume}{2019} (\bibinfo{year}{2019}).
%Type = Article
\bibitem[{To{\u{g}}a{\c{c}}ar et~al.(2020)To{\u{g}}a{\c{c}}ar, Ergen,
  C{\"o}mert, and {\"O}zyurt}]{tougaccar2020deep}
\bibinfo{author}{M.~To{\u{g}}a{\c{c}}ar}, \bibinfo{author}{B.~Ergen},
  \bibinfo{author}{Z.~C{\"o}mert}, \bibinfo{author}{F.~{\"O}zyurt},
\newblock \bibinfo{title}{A deep feature learning model for pneumonia detection
  applying a combination of mrmr feature selection and machine learning
  models},
\newblock \bibinfo{journal}{IRBM} \bibinfo{volume}{41} (\bibinfo{year}{2020})
  \bibinfo{pages}{212--222}.
%Type = Inproceedings
\bibitem[{Jakhar and Hooda(2018)}]{jakhar2018big}
\bibinfo{author}{K.~Jakhar}, \bibinfo{author}{N.~Hooda},
\newblock \bibinfo{title}{Big data deep learning framework using keras: A case
  study of pneumonia prediction},
\newblock in: \bibinfo{booktitle}{2018 4th International Conference on
  Computing Communication and Automation (ICCCA)},
  \bibinfo{organization}{IEEE}, \bibinfo{year}{2018}, pp.
  \bibinfo{pages}{1--5}.
%Type = Book
\bibitem[{Kleinbaum et~al.(2002)Kleinbaum, Dietz, Gail, Klein, and
  Klein}]{kleinbaum2002logistic}
\bibinfo{author}{D.~G. Kleinbaum}, \bibinfo{author}{K.~Dietz},
  \bibinfo{author}{M.~Gail}, \bibinfo{author}{M.~Klein},
  \bibinfo{author}{M.~Klein}, \bibinfo{title}{Logistic regression},
  \bibinfo{publisher}{Springer}, \bibinfo{year}{2002}.
%Type = Article
\bibitem[{Myles et~al.(2004)Myles, Feudale, Liu, Woody, and
  Brown}]{myles2004introduction}
\bibinfo{author}{A.~J. Myles}, \bibinfo{author}{R.~N. Feudale},
  \bibinfo{author}{Y.~Liu}, \bibinfo{author}{N.~A. Woody},
  \bibinfo{author}{S.~D. Brown},
\newblock \bibinfo{title}{An introduction to decision tree modeling},
\newblock \bibinfo{journal}{Journal of Chemometrics: A Journal of the
  Chemometrics Society} \bibinfo{volume}{18} (\bibinfo{year}{2004})
  \bibinfo{pages}{275--285}.
%Type = Article
\bibitem[{Ke et~al.(2017)Ke, Meng, Finley, Wang, Chen, Ma, Ye, and
  Liu}]{ke2017lightgbm}
\bibinfo{author}{G.~Ke}, \bibinfo{author}{Q.~Meng},
  \bibinfo{author}{T.~Finley}, \bibinfo{author}{T.~Wang},
  \bibinfo{author}{W.~Chen}, \bibinfo{author}{W.~Ma}, \bibinfo{author}{Q.~Ye},
  \bibinfo{author}{T.-Y. Liu},
\newblock \bibinfo{title}{Lightgbm: A highly efficient gradient boosting
  decision tree},
\newblock \bibinfo{journal}{Advances in neural information processing systems}
  \bibinfo{volume}{30} (\bibinfo{year}{2017}) \bibinfo{pages}{3146--3154}.
%Type = Article
\bibitem[{Pal and Foody(2010)}]{pal2010feature}
\bibinfo{author}{M.~Pal}, \bibinfo{author}{G.~M. Foody},
\newblock \bibinfo{title}{Feature selection for classification of hyperspectral
  data by svm},
\newblock \bibinfo{journal}{IEEE Transactions on Geoscience and Remote Sensing}
  \bibinfo{volume}{48} (\bibinfo{year}{2010}) \bibinfo{pages}{2297--2307}.
%Type = Inproceedings
\bibitem[{Godbole and Sarawagi(2004)}]{godbole2004discriminative}
\bibinfo{author}{S.~Godbole}, \bibinfo{author}{S.~Sarawagi},
\newblock \bibinfo{title}{Discriminative methods for multi-labeled
  classification},
\newblock in: \bibinfo{booktitle}{Pacific-Asia conference on knowledge
  discovery and data mining}, \bibinfo{organization}{Springer},
  \bibinfo{year}{2004}, pp. \bibinfo{pages}{22--30}.
%Type = Article
\bibitem[{Wang et~al.(2012)Wang, Gill, Perera, Thomson, Mant, and
  Harnden}]{wang2012clinical}
\bibinfo{author}{K.~Wang}, \bibinfo{author}{P.~Gill},
  \bibinfo{author}{R.~Perera}, \bibinfo{author}{A.~Thomson},
  \bibinfo{author}{D.~Mant}, \bibinfo{author}{A.~Harnden},
\newblock \bibinfo{title}{Clinical symptoms and signs for the diagnosis of
  mycoplasma pneumoniae in children and adolescents with community-acquired
  pneumonia},
\newblock \bibinfo{journal}{Cochrane Database of Systematic Reviews}
  (\bibinfo{year}{2012}).
%Type = Article
\bibitem[{Rhim et~al.(2019)Rhim, Kang, Yang, and Lee}]{rhim2019epidemiological}
\bibinfo{author}{J.~W. Rhim}, \bibinfo{author}{H.~M. Kang},
  \bibinfo{author}{E.~A. Yang}, \bibinfo{author}{K.~Y. Lee},
\newblock \bibinfo{title}{Epidemiological relationship between mycoplasma
  pneumoniae pneumonia and recurrent wheezing episode in children: an
  observational study at a single hospital in korea},
\newblock \bibinfo{journal}{BMJ open} \bibinfo{volume}{9}
  (\bibinfo{year}{2019}) \bibinfo{pages}{e026461}.

\end{thebibliography}





% \end{thebibliography}

\end{document}